\documentclass[sigconf]{acmart}
\renewcommand\footnotetextcopyrightpermission[1]{} % removes footnote with conference information in first column
\usepackage{multirow}
\renewcommand{\paragraph}[1]{
     \textbf{#1.} 
 }

\let\svthefootnote\thefootnote
\newcommand\freefootnote[1]{%
  \let\thefootnote\relax%
  \footnotetext{#1}%
  \let\thefootnote\svthefootnote%
}

\DeclareMathOperator*{\s}{s}
\AtBeginDocument{%
  \providecommand\BibTeX{{%
    \normalfont B\kern-0.5em{\scshape i\kern-0.25em b}\kern-0.8em\TeX}}}

\setcopyright{acmcopyright}
\copyrightyear{2023}
\acmYear{2023}
\acmDOI{XXXXXXX.XXXXXXX}
\acmSubmissionID{2277}
\setcopyright{acmcopyright}\acmConference[WWW '23]{The Web Conference}{April 30--May 04, 2023}
{Austin, Texas}
\begin{document}

% \title{Bi-directional Link: Bridging Inductive Link Predictions with Contrastive Learning of Transformers}
\title{Bi-Link: Bridging Inductive Link Predictions from Text via Contrastive Learning of Transformers and Prompts}

\author{Bohua Peng}
\affiliation{%
  \institution{Modelbest}
  \city{London}
  \country{UK}
}
\email{bp1119@ic.ac.uk}
\author{Shihao Liang}
\affiliation{%
  \institution{Tsinghua University}
  \city{Beijing}
  \country{China}
}
\email{shihaoliang0828@163.com}
\author{Mobarakol Islam}
\affiliation{%
  \institution{University College London}
  \city{London}
    \country{UK}
  }
\email{mobarakol.islam@ucl.ac.uk}

\renewcommand{\shortauthors}{Bohua Peng, Shihao Liang, Mobarakol Islam.}

\begin{abstract}
Inductive knowledge graph completion requires models to comprehend the underlying semantics and logic patterns of relations. With the advance of pretrained language models, recent research have designed transformers for link prediction tasks. However, empirical studies show that linearizing triples affects the learning of relational patterns, such as inversion and symmetry. In this paper, we propose Bi-Link, a contrastive learning framework with probabilistic syntax prompts for link predictions. Using grammatical knowledge of BERT, we efficiently search for relational prompts according to learnt syntactical patterns that generalize to large knowledge graphs. To better express symmetric relations, we design a symmetric link prediction model, establishing bidirectional linking between forward prediction and backward prediction. This bidirectional linking accommodates flexible self-ensemble strategies at test time. In our experiments, Bi-Link outperforms recent baselines on link prediction datasets (WN18RR, FB15K-237, and Wikidata5M).
Furthermore, we construct Zeshel-Ind as an in-domain inductive entity linking the environment to evaluate Bi-Link. The experimental results demonstrate that our method yields robust representations which can generalize under domain shift. Our code and dataset are publicly available at \url{https://anonymous.4open.science/r/Bi-Link-2277/.}
\end{abstract}





\maketitle
\begin{figure}[t]
\begin{center}
% \begin{overpic} 
% [width=\linewidth]
% {example-image-a}
% \end{overpic}
\includegraphics[width=1.05\linewidth]{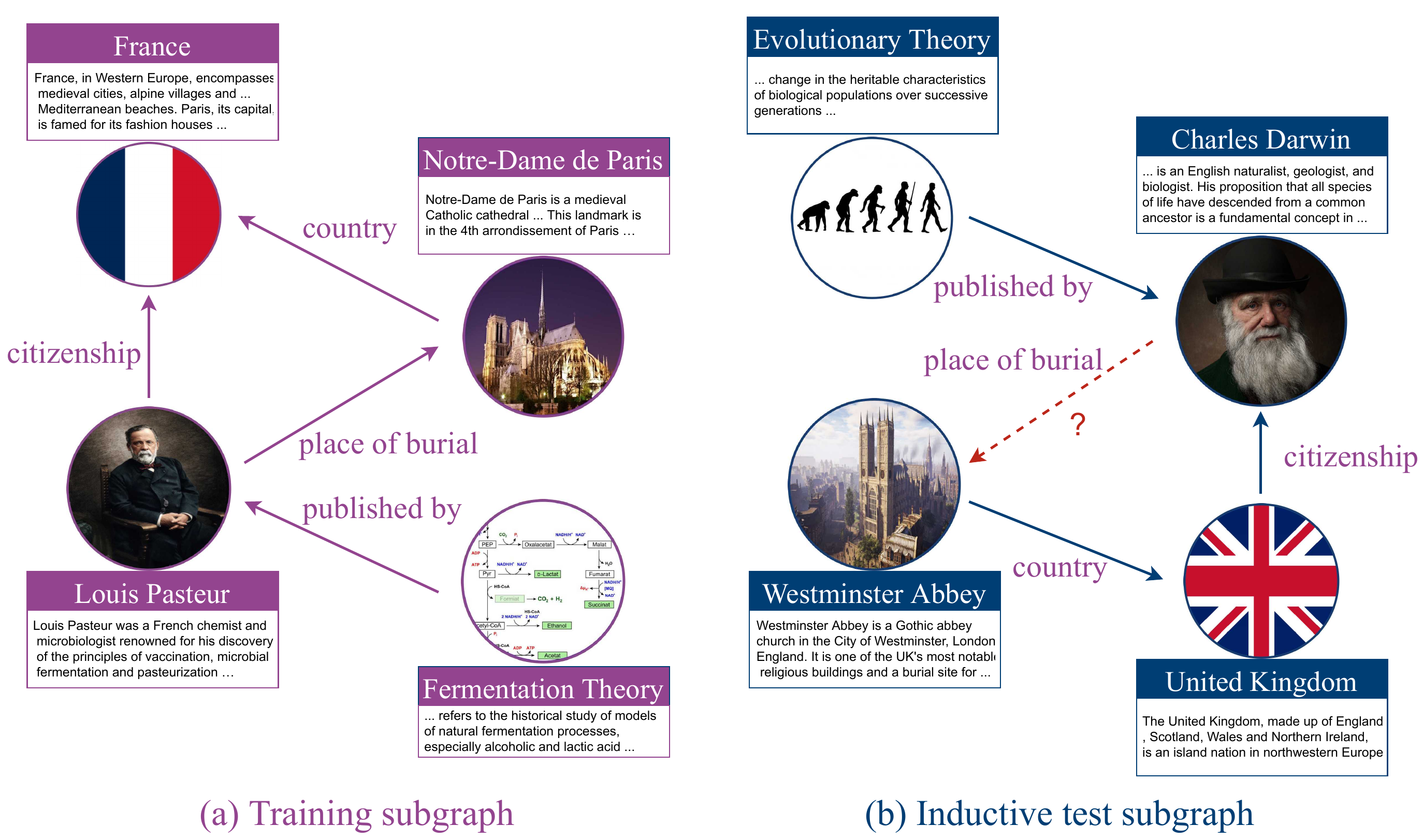}
\end{center}
\caption{
A toy example of inductive link predictions in knowledge graphs. The training subgraph and test subgraph have mutual relations but disjoint entities. Entity descriptions can help structural generalization from entities of the training subgraph (purple) to unseen entities of the test subgraph (blue). 
}
\label{fig:teaser}
\end{figure}
\section{Introduction}
Knowledge graphs are structured fact databases representing entities as nodes and relations as edges.
With open-end incoming data, automatically completing knowledge graphs is an a-billion-dollar problem for knowledge-intensive tasks, such as question answering~\cite{ma2022open} and dialogue systems~\cite{Saxena2022SequencetoSequenceKG}. As a particularly popular paradigm, TransE~\cite{bordes2013translating} initially proposes an additive model for knowledge representations. Despite its simplicity, TransE has modelled relational patterns such as inversion and composition. ComplEx~\cite{trouillon2016complex} represents symmetric relations with a real embedding space and an imaginary embedding space. With a multiplicative model, RotatE~\cite{sun2019rotate} has extended expressiveness to most general patterns, including antisymmetry and reflexiveness.

However, link prediction in the real world is an inductive learning process, as shown in Figure~\ref{fig:teaser}, where models are not only required to understand logical patterns but also reason on unseen entities.
% Predicting missing relations on existing knowledge graphs is a widely studied problem. 
% As extremely popular paradigms, TransE~\cite{bordes2013translating} and its variants have learnt expressive knowledge graph representations from entities.
To perform well in the inductive setting, models should command the relational semantics of knowledge graphs, i.e., the logical rules among the relations. For example, an intelligent model should have an induction ability to put entities in logical frames, as follows
\begin{align}\label{eq:rule}
\begin{split}
    \exists Y. (X, son\_of, Y)\wedge &(Z, daugther\_of, Y)\\
    &\rightarrow (X, sibling\_of, Z)
\end{split}
\end{align}
DKRL~\cite{xie2016representation}, a pioneering inductive link prediction method,  proposes to ground these logical rules from entity descriptions. KEPLER~\cite{wang2021kepler} incorporates the logic of triples and semantics of entity description encoded with the advance of pre-trained language models (PLM). The loss function is a linear combination of TransE’s loss and masked-language modelling loss. Unfortunately, the method converges unexpectedly slow and shows sensitivity to noise in entity descriptions. With highly efficient bi-encoders, SimKGC~\cite{wang2022simkgc} effectively compares knowledge representations using normalized contrastive loss~\cite{chen2020simple}. With very fast convergence, this simple method completes link prediction on FB15k-237 with $\sim$10min, whereas KEPLER needs $\sim$20h.

However, empirical studies show that the recent contrastive link prediction method heavily relies on the semantic match and ignores the underlying graphical structure. 
  We hypothesize flawed initialization is the culprit of poor contrastive representation learning. In other words, it is hard for PLMs to estimate the similarity between linearized relational expression and entity, e.g., comparing “X inverse sibling of” with “Z”.
  
\begin{figure}[htbp]
    \centering
    \includegraphics[width=0.92\linewidth]{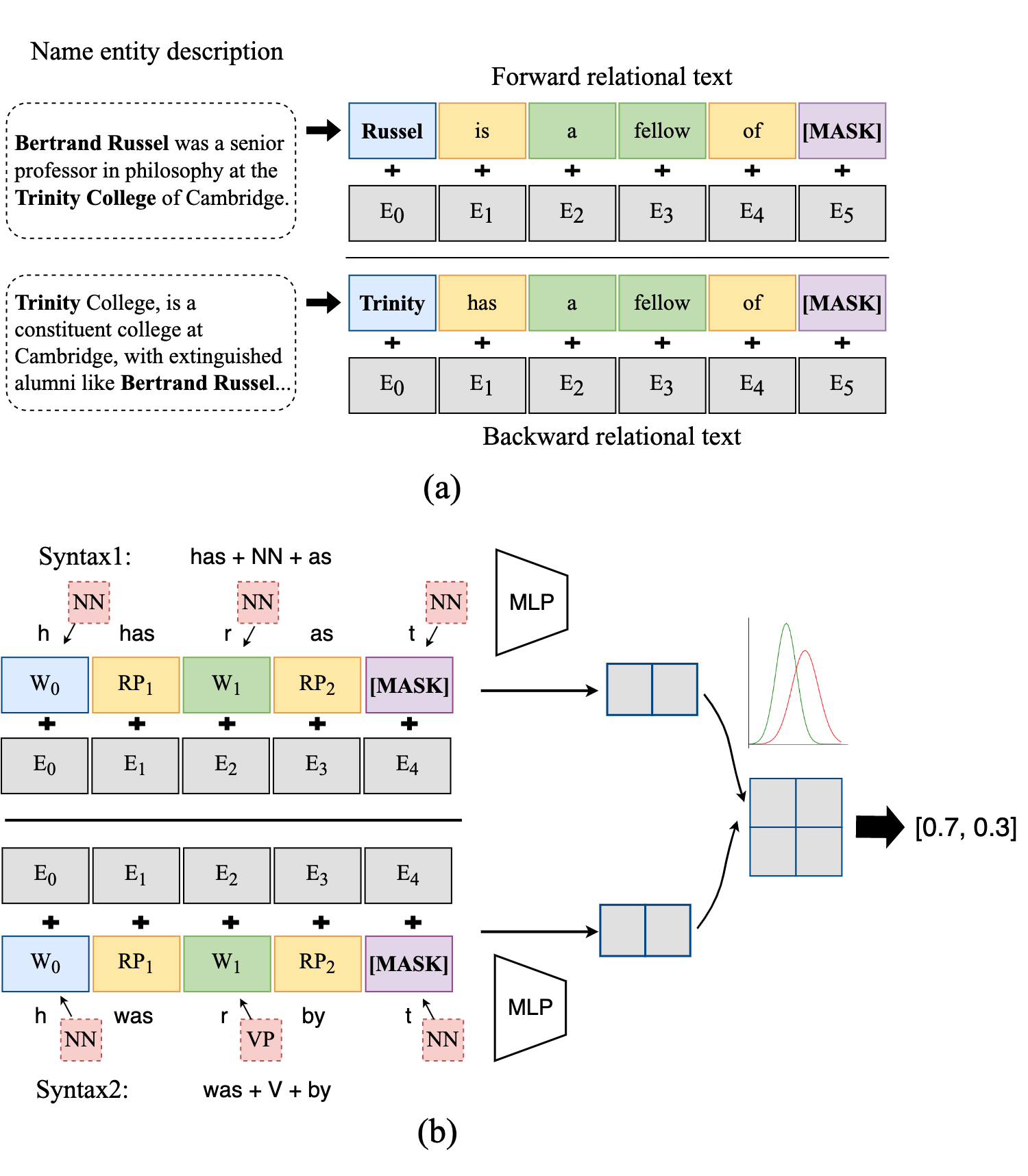}
    \caption{Probabilistic syntactic prompts. (a) Reversible prompts with syntax, NN NN NN. (b) The lightweight syntactic prompt generator with POS embedded in pretraining.}
    \label{fig:RP}
\end{figure}

This phenomenon reveals at least two problems. First, different semantics need different ways to express inversion. Motivated by recent prompt-tuning studies~\cite{lester-etal-2021-power, gao2020making}, we generate ruled-based prompts to improve the fluency of the relational expression under different types of syntax. Using syntax as the principle to find prompts not only improves interpretability but also reduces the difficulty of the prompt search, as there are not too many grammar rules in English. Specifically, we finetune PLMs on the part-of-speech tagging task to encode grammar and text into the same semantics space. Using the embeddings of PLMs, in Figure \ref{fig:RP}, we train a multi-layer perceptron to predict the syntactical patterns in a smooth low-dimensional space. Then a Gaussian mixture model indexes a pair of reversible prompts for forward and backward link predictions. We combine the prompts and unfinished edges as relational expressions to better finetune bi-encoders for inductive link predictions. The parameters of the MLP and bi-encoders are updated with the expectation maximization (EM) algorithm~\cite{li2020dividemix}.

\begin{figure*}
\begin{center}
\includegraphics[width=0.48\linewidth]{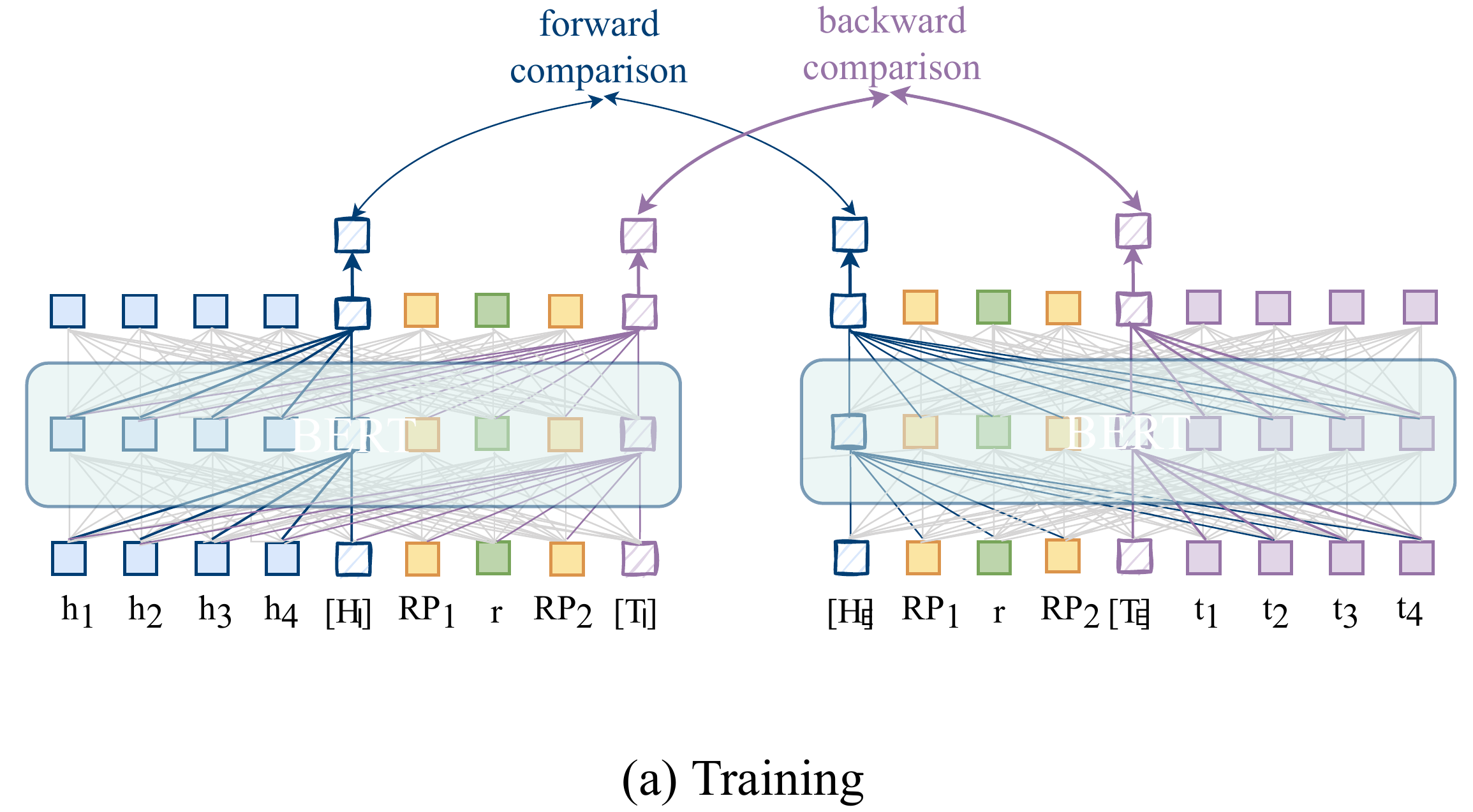}
\includegraphics[width=0.48\linewidth]{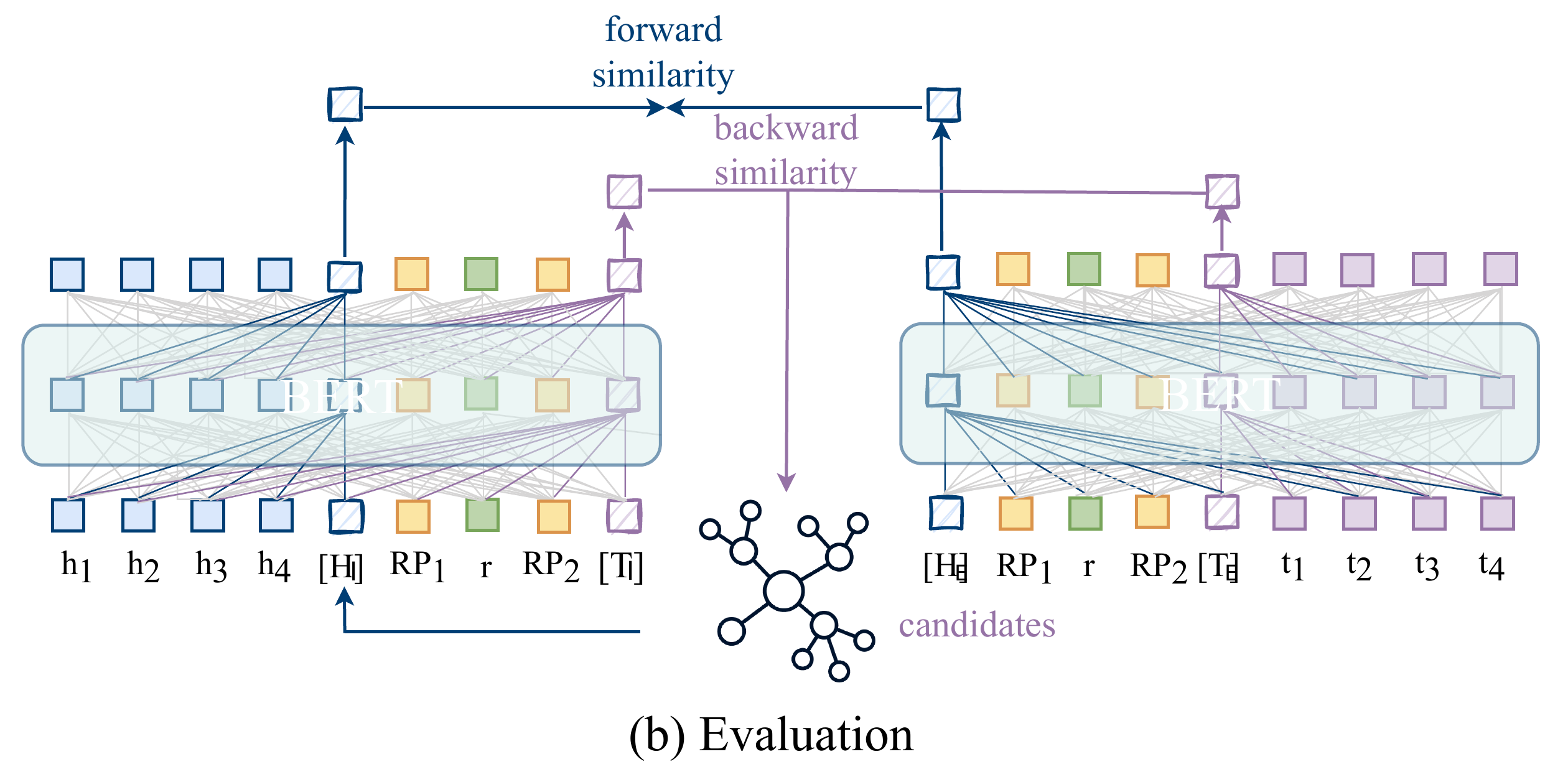}
\end{center}
\caption{
An overview of the proposed Bi-Link model, a siamese network consisting of a pair of transformer encoders. (a) In the forward prediction, a tail entity is represented with the related head entity's description (blue), rule-based prompt (yellow) and relation (green) as $[T_{f}]$. In the backward prediction, a head entity is represented with the related tail entity's description (purple), rule-based prompt (yellow) and relation (green) as $[H_{b}]$. The contrastive learning pulls the $[H]$s and $[T]$s of the same triple together while pushing the negative in-batch pairs apart. (b) During evaluation, the bidirectional linking effect allows the prediction of one direction to improve another. It also enables post-hoc entity embedding updates by averaging $[H_{f}]$ with its new neighbours $[H_{b}]$.
}
\label{fig:overview}
\end{figure*}

Another problem is that recent contrastive representation learning devastatingly affects symmetric relation modelling. Symmetry has deeply rooted foundations in neural computing~\cite{zbontar2021barlow}. To improve expressiveness on symmetry, we introduce bidirectional linking between relational bi-encoders (Figure \ref{fig:overview}), briefly known as Bi-Link. Given relational expressions, Bi-Link understands a triple from both directions, getting better comprehension for symmetric relations like “sibling”. Interestingly, the bidirectional linking learnt in training accommodates flexible self-ensembling options for test-time augmentation. Bi-Link outperforms recent baselines in our experiments on transductive and inductive link prediction tasks.

Bi-Link can be applied to knowledge-intensive tasks with minimal modification. With rich lexical and syntactical semantics, the entity linking task ranks all candidate documents to predict a link from a named entity mention in context to its referent entity document. Recently, Zeshel~\cite{logeswaran2019zero} has greatly supported the zero-shot evaluation of PLMs with an entity linking corpus beyond factual knowledge bases intensively used 
for pretrainig. However, Zeshel combines the tasks of zero-shot learning and unsupervised domain adaptation. Technically, this has caused a de facto pain to error analysis of different models and training methods. In this work, we create an in-domain zero-shot entity linking benchmark, Zeshel-Ind, with data of Fandom. To adapt our model to zero-shot entity linking, we use shared soft prompts for mention spans instead of prompts on both sides. Besides recent contrastive techniques used in SimCSE~\cite{gao2021simcse} and SimKGC~\cite{wang2022simkgc}, we opt to share negative candidates retrieved by BM25~\cite{robertson2009probabilistic} among in-batch samples. Our methods have achieved competitive results on both Zeshel-Ind and Zeshel. Our experiments also show interesting behaviour of contrastive learning toward domain shift.\\
In summary, our major contributions are threefold:
1.) We propose a probabilistic syntactic prompt method to verbalize the unfinished edge into natural relational expressions with a generalizable lightweight model.
2.)We design symmetric relational encoders, Bi-Link, for text-based knowledge graph link predictions, and adapt it to entity linking tasks.
3.)We build a new open-source benchmark, Zeshel-Ind as a fully-inductive reflection of Zeshel for in-domain zero-shot performance evaluation. We extensively validate Bi-Link with several recent baselines on Zeshel and Zeshel-Ind.
\section{Related work}
% Link predictionRecent years have witnessed increasing interest in knowledge graph link prediction, with various knowledge representations devised.
% TransE~\cite{bordes2013translating} is the earliest translation-based embedding model. However, these models is reported to have low expressive. The second group of multiplicative models includes ComplEx~\cite{trouillon2016complex} and RotatE~\cite{sun2019rotate}, encoding more patterns including symmetry and reflexiveness. While their intrinsic associations are neglected, which is insufficient for capturing deeper semantics for better embedding. Neural network models such as ConvE~\cite{dettmers2018convolutional} are able to consider the type of entity or relation, temporal information, path information and substructure information, which is deeper than the semantic-based models. 
% With the advent of the pre-training paradigm, it is possible to learn graph representations from the knowledge of PLMs with textual descriptions of entities.
\paragraph{Contrastive knowledge representation}Contrastive knowledge representation
Inspired by NCE~\cite{gutmann2010noise} principle, CPC~\cite{oord2018representation} and SimCLR~\cite{chen2020simple} are particularly popular contrastive learning paradigms that learn robust representation with noisy negative samples.
For semantic textual similarity tasks (STS), SimCSE~\cite{gao2021simcse} significantly simplifies previous contrastive sentence embedding methods using bi-transformers. PromptBERT~\cite{jiang2022promptbert} has further improved the results with template denoising. However, empirical studies~\cite{wang2021gpl} show that SimCSE is not helpful for entity linking. Using bi-transformers to contrast document representations is still an ongoing research direction. Finding proper negative samples is crucial for contrastive learning~\cite{wang2022simkgc}. GPL~\cite{wang2021gpl} automatically find high-quality negative samples with a pipeline, including T5~\cite{raffel2020exploring}, dense retriever~\cite{robertson2009probabilistic} and a cross-encoder~\cite{humeau2019poly}. In this work, we generalize a lightweight syntactic prompt generator learnt on a subgraph to a large knowledge graph. The prompts improve the quality of negative samples by transferring unfinished edges to approximate relational expressions.\\
\paragraph{Retrieval with pretrained transformers}
Bi-encoders independently map queries and documents to a shared semantic space to efficiently compute their similarity scores. By contrast, cross-encoders~\cite{yao2019kg} broadcast a query and concatenate it with all possible documents, predicting relevance scores with cross-attention between the query and documents. Previous work~\cite{humeau2019poly} has shown that cross-encoders can produce more robust representations and achieve better results. But the cumbersome computational overhead harshly increases the inference time. To address this problem, ColBERT~\cite{khattab2020colbert} and TwinBERT~\cite{lu2020twinbert} concurrently propose hybrid networks consisting of bi-encoders and cross-attention layers. As a result, these works have facilitated real-world search engines by significantly reducing the computational burden while retaining the performance. On the flip side, the late-stage interaction requires additional training data and strategy. Our work is different from the previous work because our prompt-based bi-encoders can be useful when queries are fragmented words.
% BLINK~\cite{wu2019scalable} simply retrieves candidates by a bi-encoder. Then each candidate is re-ranked with a cross-encoder.
% MuVER~\cite{ma2021muver} successfully constructs multi-view representations for entity descriptions. These methods perform great in entity retrieval, but do not capture the symmetric information within the descriptions.
\section{Methods}
Given a directed knowledge graph $\mathcal{G}=(V, E, R)$ with $|V|$ entities, $|E|$ observed edges, and $|R|$ relation types, our transductive link prediction task is to infer a missing edge $e$ described as $(h, r, ?)$  or its inverse version $e_{inv}$ described as $(?, r, t)$. As a logical operator, a triple $(h, r, t)$ consists of a head entity, relation and a tail entity. 
In this work, we present Bi-Link, a symmetric framework that learns knowledge representations from entity description and reversible relational text. As the example shown in Figure \ref{fig:RP}, we enrich $\mathcal{G} $ with inverse relational textual information phrased by probabilistic rule-based prompt expressions. Then we encode relational information with a siamese network, Bi-Link (Figure \ref{fig:overview}). The grammatical module and the Bi-Link network are updated with EM algorithm. We use this general framework for transductive link prediction, fully-inductive link prediction~\cite{teru2020inductive}, inductive named entity linking.
An overview of the propose framework is presented in Figure \ref{fig:RP} and Figure \ref{fig:overview} with each step explained in more detail below.

\subsection{Probabilistic rule-based prompts}\label{SRBP} 
To reduce the semantic gap between bi-encoders, we introduce probabilistic rule-based prompts (RP) to utilize the implicit knowledge in PLMs. 
To obtain grammatical information from contextualized embeddings, we follow the procedure of RoBERTa~\cite{liu2019roberta} to finetune BERT~\cite{devlin2018bert} on a well-known part of tagging dataset, PennTreeBank~\cite{marcus-etal-1993-building}. We use this finetuned BERT as backbone for knowledge graph link prediction. 
Given a named entity description $x=[w_{1},w_{2},...w_{n}]$ and one of the entity's outflow edge $e_{i}=(h^{i}, r^{i}, [MASK])$, as shown in Figure \ref{fig:RP}, we iteratively combine the edge with all possible prompts $\mathcal{T}_{forward}$ in our grammar rule base. The size of $\mathcal{T}_{forward}$ is m. After tokenization, we feed the tokens and positions into the BERT embedding layer and take the averaged embeddings. These relational embeddings then go through a MLP, projected as low dimensional representations $x^i$. We then compute the responsibility of Gaussian mixture models and apply a label sharpening, allowing top predictions to update. The probability of selecting a template with the edge can be formally written as
\begin{align}\label{rbp}
    P(\tau|e)&=P(\tau|h, r)\\
    &=\prod_{e\in D}\mathcal{N}(g_{\theta}(h, \tau, r)|\mu,\Sigma)
\end{align}
where $\tau$ denotes a template corresponding to a syntactical pattern. $g$, a shallow encoder parameterized with $\theta$, encodes the embeddings of an incomplete edge and the prompt of last iteration into   a lower dimensional space. $\mu$ and $\Sigma$ are the parameters of the Gaussian mixture model. The cluster number $k$ is a hyperparameter.\\
The parameters of this syntactical prompt generator and bi-encoder are updated the EM algorithm, and the M step can be written as follows,
\begin{align}
    z_{i}&=g_{\theta}(\tau, e)\\
    u_k&=\frac{\sum_{i=1}^N r_{i k} z_{i}^{max}}{N}\\
    \Sigma_k&=\sum_{i=1}^N r_{i k}\left(z_{i}^{max}-u_k\right) \cdot\left(z_{i}^{max}-u_k\right)^T
\end{align}
where $u_k$ and $\Sigma_k$ are the mean and covariance of the k th Gaussian component, $r_{i k}$ is the responsibility, $z_{i}^{max}$ is the most probable prompt-based relational expression. Additionally, the parameters of GMMs are initialized with the statistics of a manually labelled subset of the validation set. Annotators have combined these triples with most suitable prompts based on the frequency of expressions in English-Corpora\footnote{https://www.english-corpora.org}. With the most probable syntactical prompts, we verbalize the outflow edge and inflow edge as relational expressions, $v(e_{hr})$ and $v(e_{rt})$ in the input space. 
% In the following sections, we will demonstrate rule-based prompts (RP) benefits contrastive learning by improving the quality of contrastive examples.
\subsection{KG link prediction}
In Figure \ref{fig:overview}, Bi-Link uses bi-transformers to score the similarity between a relational expression and an entity, which is a ranking problem in this paper. In the forward link prediction, Bi-Link models the probability of a tail entity given the head entity's relational expression as $P(t|h, RP_1, r, RP_2)$. Similarly, the backward probability can be represented as $P(h|RP_1, r, RP_2, t)$. This model can be used for both transductive learning~\cite{bordes2013translating} and inductive learning~\cite{teru2020inductive} tasks, and the contextualized relational representations can be written as,
\begin{align}\label{representation_eqs}
T_f&=PLM(v(e_{hr}), d_h),\quad T_b=PLM(d_t)\\
H_b&=PLM(v(e_{rt}), d_t),\quad H_f=PLM(d_h),\\
h_\tau&=PLM(\tau).
\end{align}
where PLM is a pretrained language model that encodes the outflow edge $v(e_{hr})$ and the head entity's description $d_h$ as the forward relational representation $T_f$. Similarly, in the backward link prediction, the PLM encodes the inflow edge $v(e_{rt}$ and the tail entity's description as a backward relational representation $H_b$. With attention mask, the head and tail entity descriptions are encoded as $H_f$ and $T_b$. In Figure~\ref{fig:overview}, these representations are output from the places of relational markers. $h_{\tau}$ is the representation of the rule-based prompt $\tau$ for template denoising~\cite{jiang2022promptbert}. 

We learn the embeddings for link predictions by pulling the textual relational embeddings of the forward prediction and the backward prediction together, with the training objective written as,
\begin{align}
    \max \sum_{i, j \in \mathcal{E}}\s(H_{f i}, \hat{H}_{b i})+&\s(T_{b i}, \hat{T}_{f i}) -\s_{i \neq j}(\hat{H}_{b i}, \hat{T}_{f i})\label{eq:CL_objective},\\
\s\left(x_i, x_j\right)&=\frac{x_i^{\top} x_j}{\left\|x_i\right\| \cdot\left\|x_j\right\| / t},\\
\hat{H}_{b i}&=H_{b i}-h_{\tau}. \label{eq:denoising}
\end{align}
where $\mathcal{E}$ is the entity set and $\s$ is the scaled cosine similarity between embeddings. The third term of Equation \ref{eq:CL_objective} penalizes naive matching based on prompt patterns. $t$ is the temperature parameter. $\hat{T}_{f i}$ and $\hat{H}_{b i}$ are denoised templates. Initially proposed in PromptBERT~\cite{jiang2022promptbert}, template denoising improves contrastive sentence embedding by removing the syntactical bias of prompts.

In practice, we optimize the training objective shown in Equation \ref{eq:CL_objective} with the following contrastive loss function:
\begin{align}
        \mathcal{L}_{1} &=\frac{1}{|\mathcal{T}|} \sum_{i \in\mathcal{T}}\left[\log e^{\s(T_{b i}, \hat{T}_{f i})}-\log\sum_{j \in \mathcal{B}_{\textbackslash i}} e^{\s(T_{b i},\hat{T}_{f j})}\right], \\
\mathcal{L}_{2} &=\frac{1}{|\mathcal{H}|} \sum_{i \in \mathcal{H}}\left[\log e^{s(\hat{H}_{f i}, H_{b i})}-\log \sum_{j \in \mathcal{B}_{\textbackslash i}} e^{s(\hat{H}_{f i}, H_{b j})}\right],\\
\mathcal{L}_{3}&=-\log\sum_{i\neq j}e^{s(\hat{H}_{b i}, \hat{T}_{f j})},\\
\mathcal{L} &= \mathcal{L}_{1}+\mathcal{L}_{2}+\beta\cdot\mathcal{L}_{3}.
\end{align}
where $\mathcal{H}$ and $\mathcal{T}$ are textual expression sets for head and tail entities. $\mathcal{B}$ is the set of the current batch, and $\beta$ is the bidirectional linking coefficient.

At the test stage, to predict a missing inflow edge $(?, r, t)$, we compute a bidirectional similarity score between the verbalized relational expression $(v(e_{rt}), d_t)$ and all entity descriptions $d_h$. In Figure \ref{fig:overview}, Bi-Link augments test performance with self-ensembling, formally written as,
\begin{align}
    \mathrm{sp1}(c, r, t)&=\s(\hat{H}_{b}, H_{f}),\\
    \mathrm{sp2}(c, r, t)&=\s(T_{b}, \hat{T}_{f}),\\
    \mathrm{score}(h, r, t) & = \mathrm{sp1} + \mathrm{w(c)}\cdot \mathrm{sp2}
\end{align}
where $\mathrm{sp1}$ and $\mathrm{sp2}$ are the same text sequence scored from the two perspectives. $c \in \mathcal{C}$ is a candidate set suggested by $\mathrm{sp1}$. $\mathrm{w(c)}$ is a flexible weighting function. For simplicity, we follow the filtered setting of TransE. Note that there is no conflict between modelling inversed patterns and symmetric patterns because tail entities can take the place of head entities with reversible rule-based prompts mentioned in Section\ref{SRBP}.
\subsection{Inductive entity linking}
When a new group of entities come into the database, they are never really mentioned in the previous records of the current society. For new posts from this timestamp, we want to link newly appearing mentions to the new entities while disambiguating them from the old. For example, after some new fictional figures are released, the administrators want to remove some disturbing new mentions to sensitive topics that can easily breed social hatred. Therefore, we give a formal definition for inductive entity linking with the following definition.\\
\textbf{Definition} Given mention sets $\mathcal{M}$, entity sets $\mathcal{E}$, candidate sets $\mathcal{C}$,and a specific text domain $\mathcal{D}$, inductive entity linking satisfies $\forall c\in\mathcal{C}, c\subseteq\mathcal{M}_{test} \vee c \subseteq\mathcal{E}_{train}$, where $\mathcal{M}_{train}\subseteq \mathcal{E}_{train}\subseteq \mathcal{D}$, $\mathcal{M}_{test}\subseteq \mathcal{E}_{test}\subseteq \mathcal{D}$. 

With no explicit relational structures, inductive entity linking is a knowledge-intensive information retrieval task that requires semantics to disambiguate in-domain named entity mentions. In the following experiments, we compare the zero-shot performance of Bi-Link with cross-encoder baselines under the defined inductive setting and another cross-domain setting.
\begin{table*}
\caption{Statistics of benchmarks. The numbers of vertices (entities), relations and triples are denoted as $\left | V \right |$, $\left | R \right |$ and \# respectively. In the inductive settings, training subgraphs and evaluation subgraphs are disjoint, resulting in sparse graphs.}\label{datasets}
\begin{tabular}{|c|cc|cc|cc|cc|}
\hline 
\multirow{2}{*}{Dataset}& \multicolumn{2}{c|}{WN18RR} & \multicolumn{2}{c|}{FB15k-237} & \multicolumn{2}{c|}{Wikidata5M}& \multicolumn{2}{c|}{Zeshel}\\
\cline{2-9}
&Transductive&Inductive&Transductive&Inductive&Transductive&Inductive&Cross-domain&Inductive\\
\hline 
$\left | V \right |+\left | R \right |$ & 40,943+11& 7553+9 & 14,541+237&6683+219& 4, 594k+822& 4, 579k+822&492,321&477,883\\
\#training & 86,835 & 7,940 & 272,115 & 27,203 & 20,614k&20,496k&59232&52812\\
\#validation & 3,034 & 1,394 & 17, 535 & 1,416 & 5,163 &6,699&10,000 &13206\\
\#test & 3,134 & 1,429 & 20, 466 & 1,424 & 5,163& 6,894&10,000&13193\\
\hline
\end{tabular}
\end{table*}
\begin{table*}
\caption{Main results of transductive link predictions on FB15k-237 and WN18RR.}\label{transductive}
\begin{tabular}{l cccc cccc}
\toprule
\multirow{2}{*}{}&\multicolumn{4}{c}{FB15k-237}&\multicolumn{4}{c}{WN18RR}\\
\cmidrule{2-5}\cmidrule{6-9}
&MRR&H@1&H@3&H@10&MRR&H@1&H@3&H@10\\
\midrule
% \multicolumn{9}{l}{\itshape embedding-based methods}\\
TransE~\cite{bordes2013translating}&27.9&19.8&37.6&44.1&22.3&1.3&40.1&53.1\\
ComplEx~\cite{trouillon2016complex}&27.1&18.4&30.1&44.7&44.6&41.0&45.8&50.2\\
RotatE~\cite{sun2019rotate} &33.8&24.1&37.5&53.3&47.6& 42.8&49.2&57.1\\
KG-BERT~\cite{yao2019kg}&24.2&14.8&26.9&43.0&21.9&9.5&24.3&49.7\\
StAR~\cite{wang2021structure}&29.6&20.5&32.2&48.2&40.1&24.3&49.1&70.9\\
TuckER~\cite{balazevic-etal-2019-tucker}&\textbf{35.8}&\textbf{26.6}&\textbf{39.4}&\textbf{54.4}&47.0&44.3&48.2&52.6\\
SimKGC~\cite{wang2022simkgc}&33.6&24.9&36.2&51.1&66.6&58.7&71.7&80.0\\
\midrule
Bi-Link&33.9&26.4&38.3&52.6&\underline{69.2}&\underline{59.6}&\underline{73.5}&\underline{81.9}\\
Bi-Link$_{RP}$&\underline{34.3}&\underline{26.5}&\underline{38.6}&\underline{53.8}&\textbf{71.8}&\textbf{60.3}&\textbf{76.1}&\textbf{84.4}\\
\bottomrule
\end{tabular}
\end{table*}
\section{Experiments}
We have designed experiments to evaluate our proposed methods from the following aspects:
\begin{itemize}
    \item \textbf{Transductive link prediction.} In the transductive knowledge graph completion task, triples are split into training, validation and test sets, but entities are shared across different sets. This setting requires the model to comprehend relational syntactic frames from the training data and predict missing relations by putting reasonable entities into the frames. 
    \item \textbf{Inductive relation prediction.} In the fully inductive setting proposed in GraIL~\cite{teru2020inductive}, relations are shared across training, validation and test triples, but entities are disjoint. Specifically, an inductive knowledge graph dataset consists of a pair of graphs, training subgraph and inductive test subgraph. The two graphs have mutual relations but disjoint entities. Composing new entities and relations, models must learn entity-independent relational semantics to complete knowledge graphs.
    \item \textbf{Inductive entity linking.} The goal of this inductive setting is to predict links between mentions and their referent entities. Mentions and entities are from a remixed domain and disjointly split into training, validation and test sets. To disambiguate an entity's mention, models should learn the underlying semantics of mentions in their context. These semantics, on the other hand, should reflect distinction between referent entities and similar candidates.
    \item \textbf{Cross domain inductive entity linking.} Data splits are on domain level where some domains are specified as training domains and the rest are evaluation domains. The cross domain inductive setting is more difficult than the inductive setting because there are domain shifts between mention context and entity description.
\end{itemize}
\subsection{Datasets}
The statistics of the datasets used in our experiments are shown in Table. The WN18RR~\cite{bordes2013translating}, FB15k-237~\cite{toutanova2015representing} and Wikidata5M~\cite{wang2021kepler} are standard benchmarks for link predictions as available in the literature. These datasets are de facto built for transductive link predictions by their original designs. 
To evaluate inductive link prediction, we employ the data splits of GraIL~\cite{teru2020inductive} where the entities of training samples and test samples are from disjoint subgraphs.
For entity linking, we consider Zeshel~\cite{logeswaran2019zero} as the cross-domain entity linking benchmark. This dataset holds community-written encyclopedias specializing in different topics, such as a fictional world or a parallel universe. Entities and mentions are from 16 topics, with 8 for training and each 4 for validation and test. Different topics have no mutual mentions or entities. Zeshel is a superb benchmark for evaluating inductive learning methods. Specifically, entities commonly presented as subjects may be present as objects in Fandom.

To distinguish inductive entity linking with domain adaptation,
we create Zeshel-Ind, an in-domain zero-shot benchmark with disjoint entity-mention splits. In detail, we first create an entity-mention set in each domain and split these sets into training, validation and test subsets by uniformly sampling mentions without replacement. Hence the subsets have disjoint entities. In our experiments, we retrieve top 64 candidates with Apache Lucene~\cite{lucenescoring}. We remove samples without any coarse search candidate. These hard positive pairs only constitute 0.1\% of the entire population and are also not used in the evaluation of ~\cite{logeswaran2019zero}. For samples with less than 64 candidates, we add randomly sampled in-domain documents until the length of the candidate list is 64. These samples constitute 23\% of the dataset. 
We remove entity names from candidate documents to avoid potential label leakage, and to penalize naive text matching. Finally, we correspondingly merge the subsets of different domains into three data splits, namely training, validation and test sets.
\begin{table*}[htbp]
\caption{Main results for entity linking on cross-domain Zeshel and inductive Zeshel.}\label{zeshel}
\begin{tabular}{l cccc cccc}
\toprule
\multirow{2}{*}{}&\multicolumn{4}{c}{Zeshel}&\multicolumn{4}{c}{Zeshel-Ind}\\
\cmidrule{2-5}\cmidrule{6-9}
&MRR&H@1&H@3&H@10&MRR&H@1&H@3&H@10\\
\midrule
% \multicolumn{9}{l}{\itshape embedding-based methods}\\
% BM25~\cite{robertson2009probabilistic}&&&&\\
BERT~\cite{devlin2018bert} &22.5&10.0&23.6&49.8&26.2& 12.7&28.9&56.1\\
SpanBERT~\cite{joshi2020spanbert} & 19.9&8.2&21.9&42.6&23.3&11.3&25.5&46.9\\
KG-BERT~\cite{yao2019kg}&72.8&\underline{60.6}&82.9&92.2&77.9&66.4&86.3&94.5\\
BLINK~\cite{wu2019scalable}&65.2&53.4&66.9&74.8&72.5&61.3&75.4&90.1\\
MuVER~\cite{ma2021muver}&55.4&45.4&70.4&81.7&78.3&67.0&86.2&95.4\\
% ColBERT~\cite{khattab2020colbert}&-&-&-&-&-&-&-&-\\
\midrule
Bi-Link&\underline{73.0}&\textbf{60.9}&\underline{82.8}&\underline{93.1}&\underline{79.8}&\underline{69.7}&\underline{87.9}&\underline{95.9}\\
Bi-Link$_{R-Drop}$&\textbf{73.3}&\textbf{60.9}&\textbf{83.4}&\textbf{93.5}&\textbf{79.9}&\textbf{70.2}&\textbf{88.4}&\textbf{96.1}\\
\bottomrule
\end{tabular}
\end{table*}
\subsection{Implementation details}\paragraph{Knowledge graph link prediction}In the transductive link prediction setting, we run experiments on three KGs of different sizes with the number of nodes varying from $\sim$15k to $\sim$5M. As for shallow embedding models, we use the codes of TransE~\cite{bordes2013translating}, ComplEx~\cite{trouillon2016complex} and RotatE~\cite{sun2019rotate} implemented by GraphVite\footnote{https://graphvite.io/docs/latest/benchmark}~\cite{zhu2019graphvite} and report the results. We re-implement KG-BERT~\cite{yao2019kg}, and reuse the official implementations of KEPLER~\cite{wang2021kepler} and SimKGC~\cite{wang2022simkgc} as baselines. 

\begin{figure*}[ht]
\centering
\includegraphics[width=0.33\linewidth]{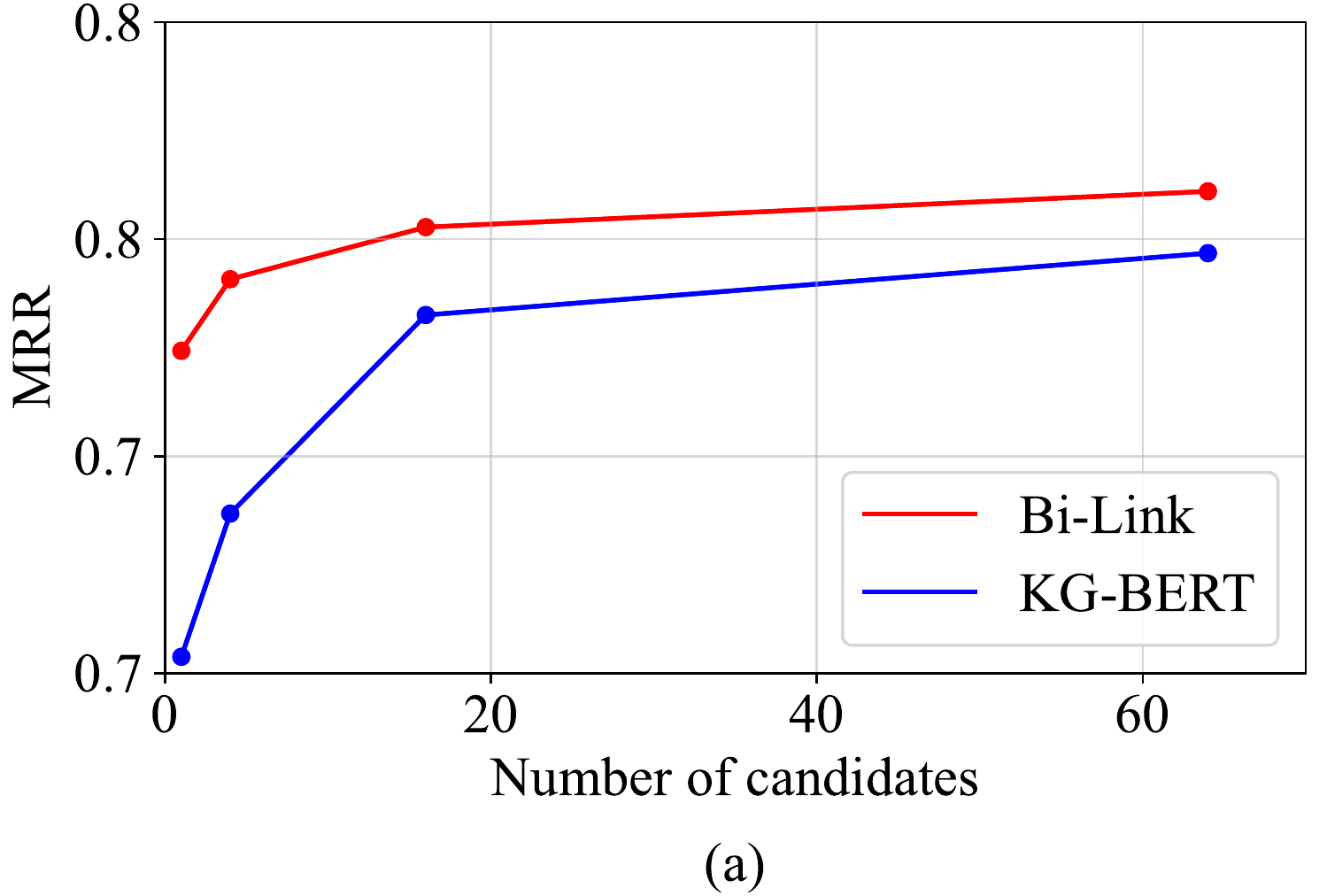}
\includegraphics[width=0.33\linewidth]{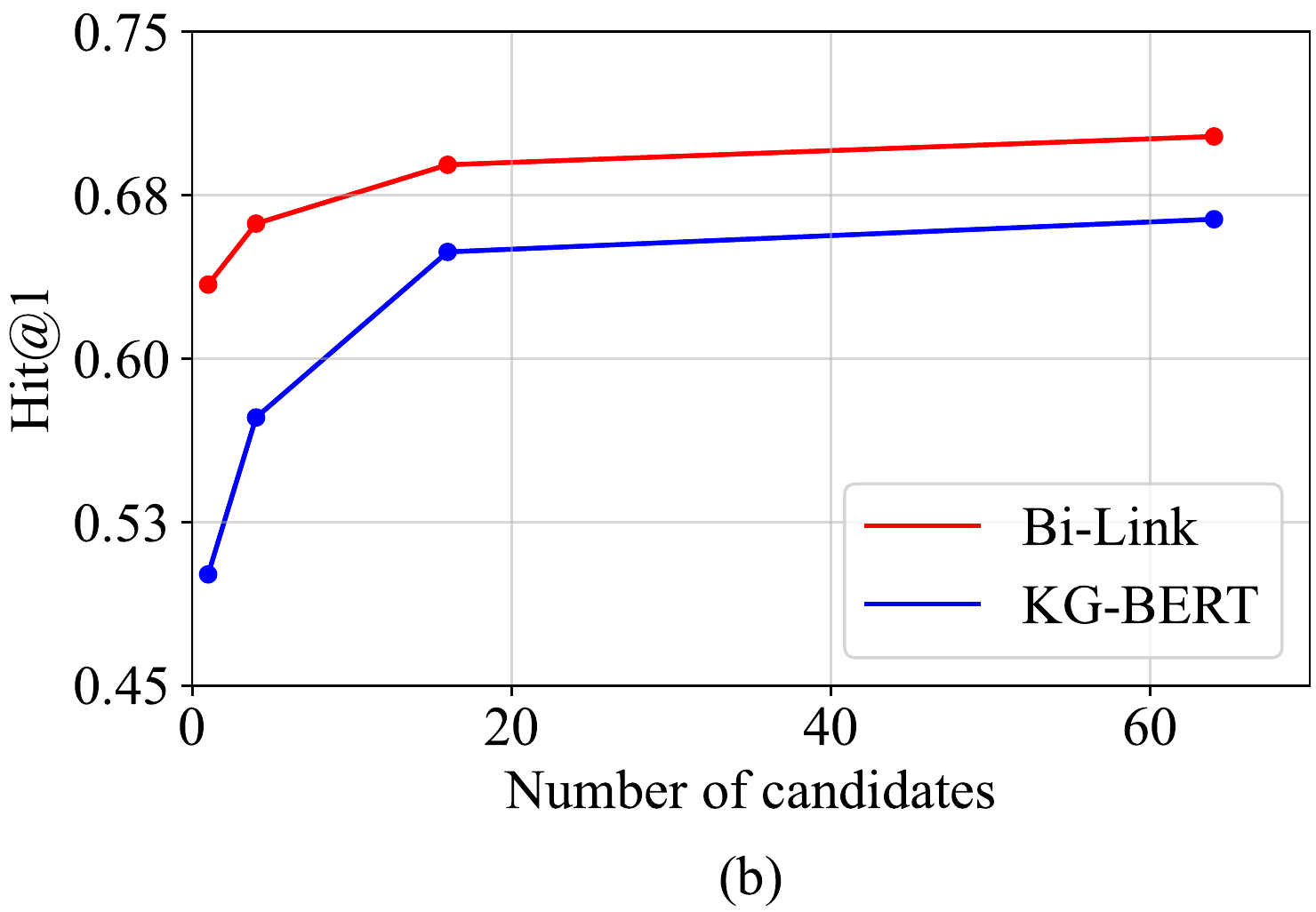}
\includegraphics[width=0.33\linewidth]{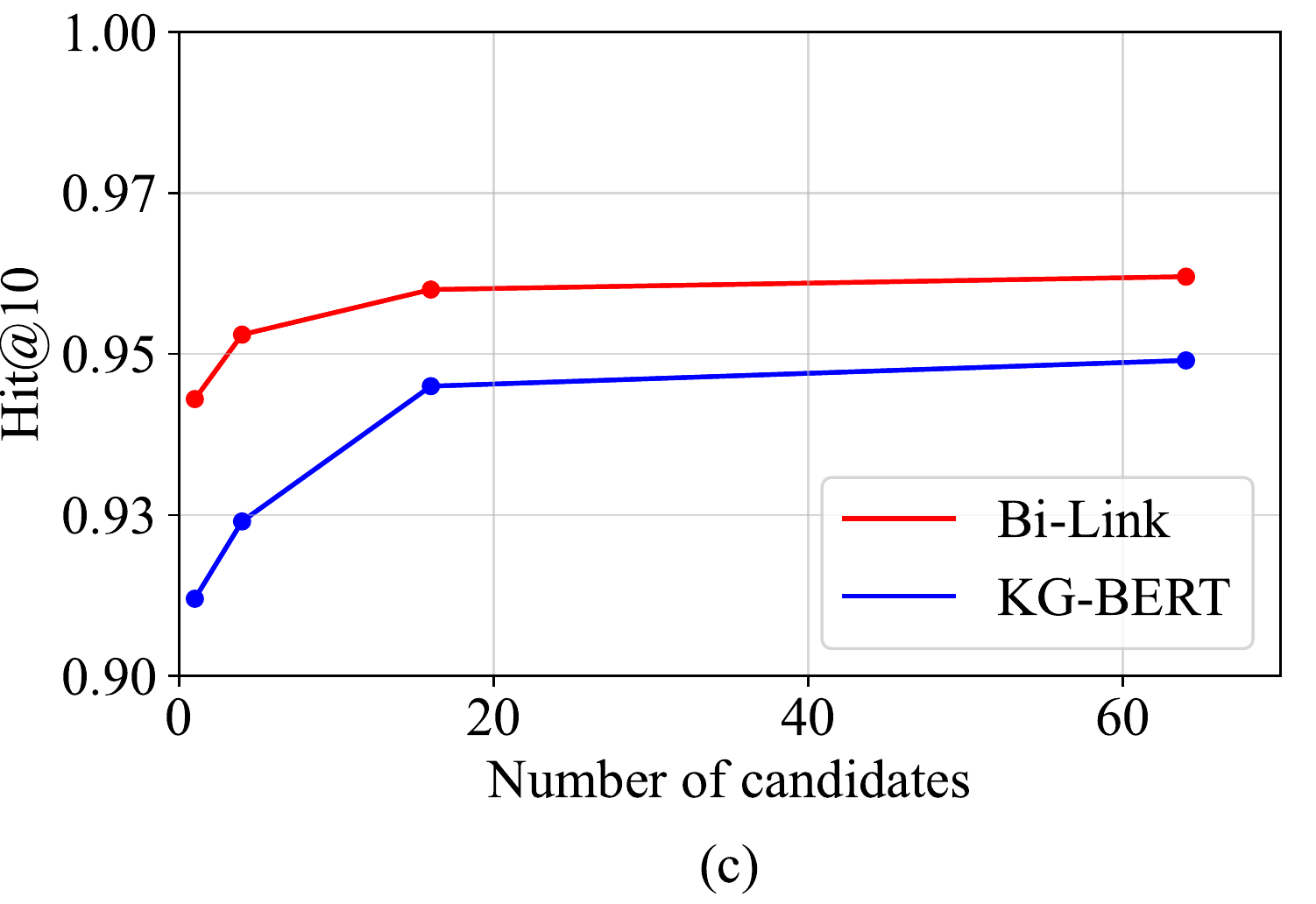}\quad
\includegraphics[width=0.33\linewidth]{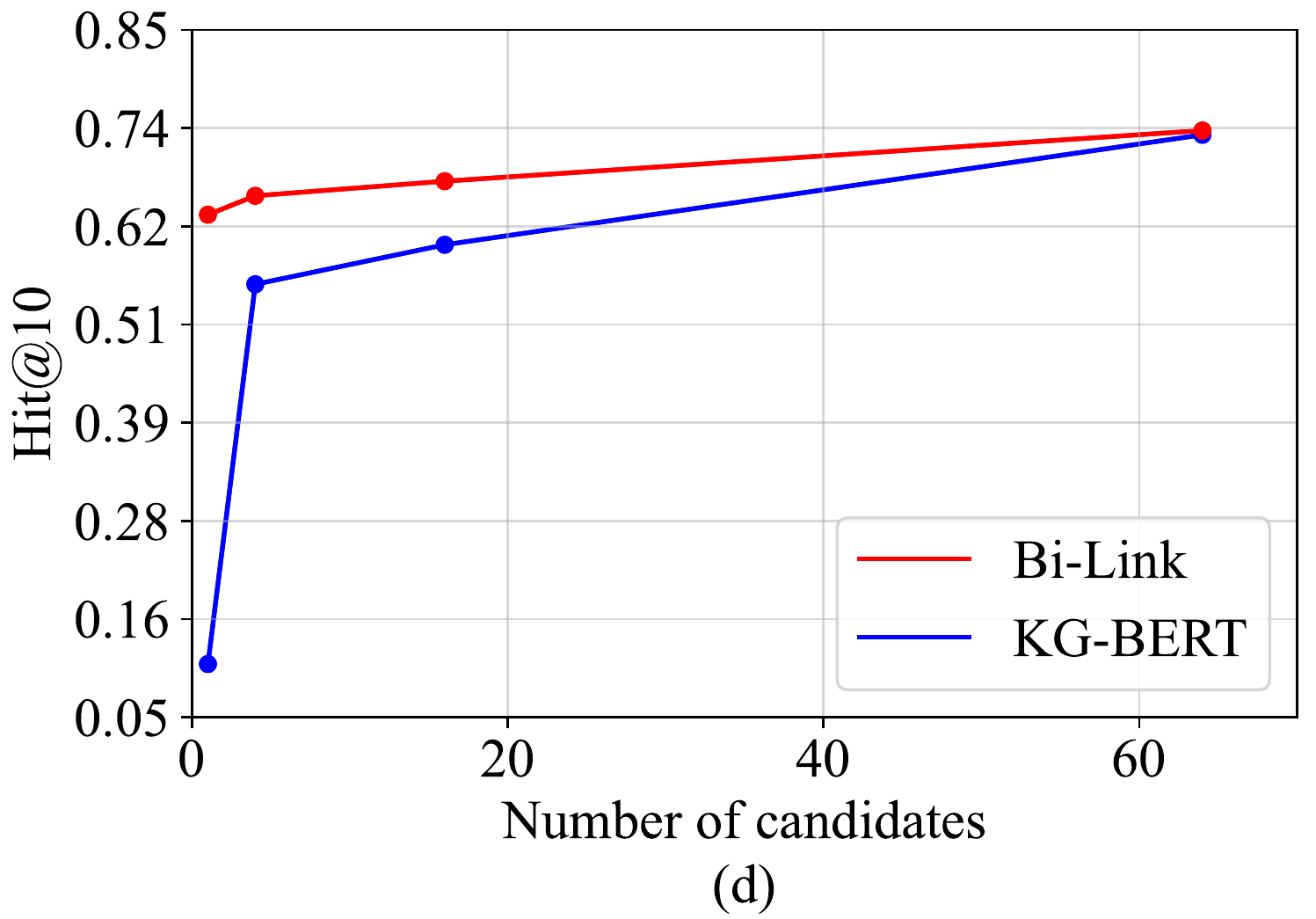}
\includegraphics[width=0.33\linewidth]{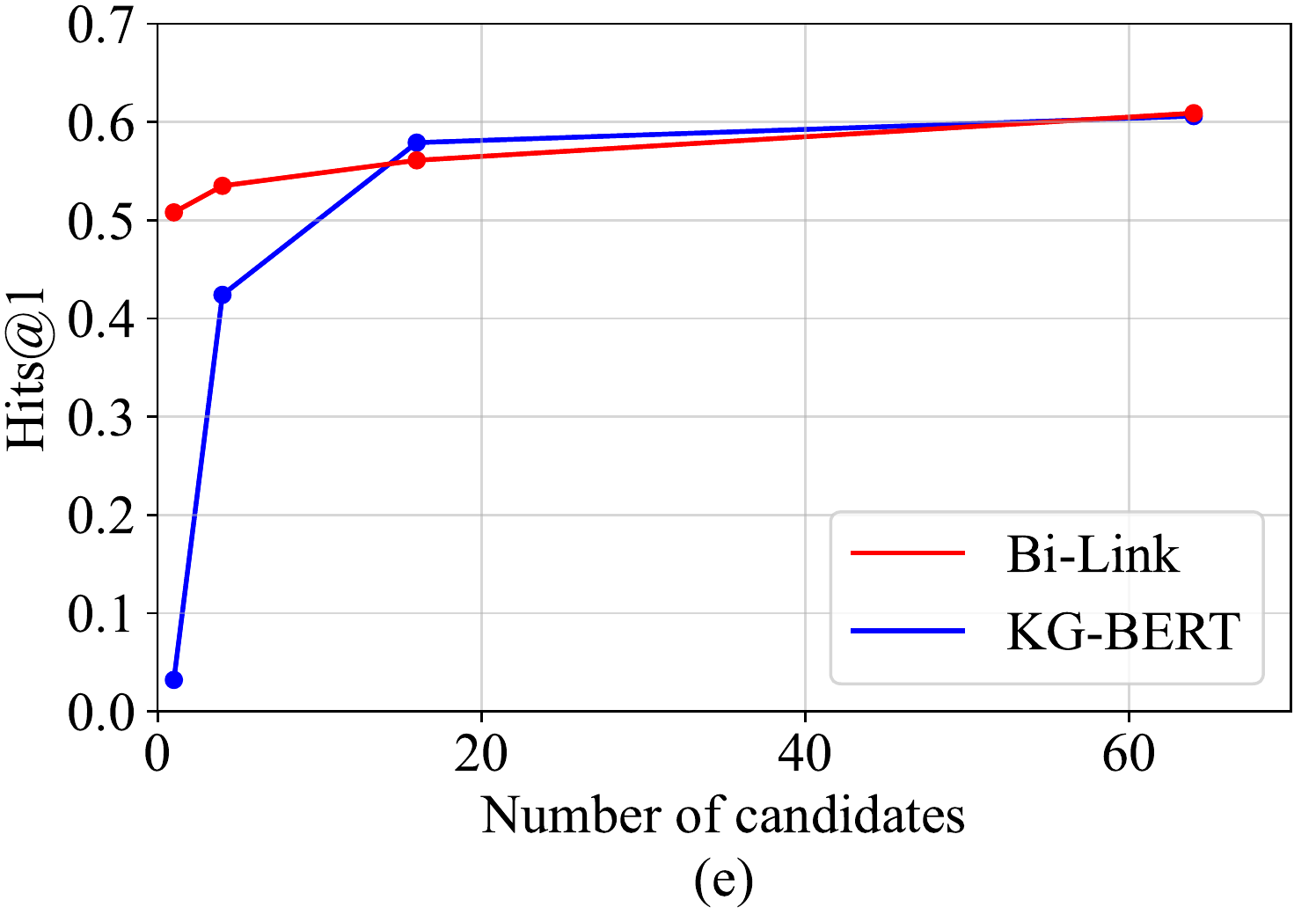}
\includegraphics[width=0.33\linewidth]{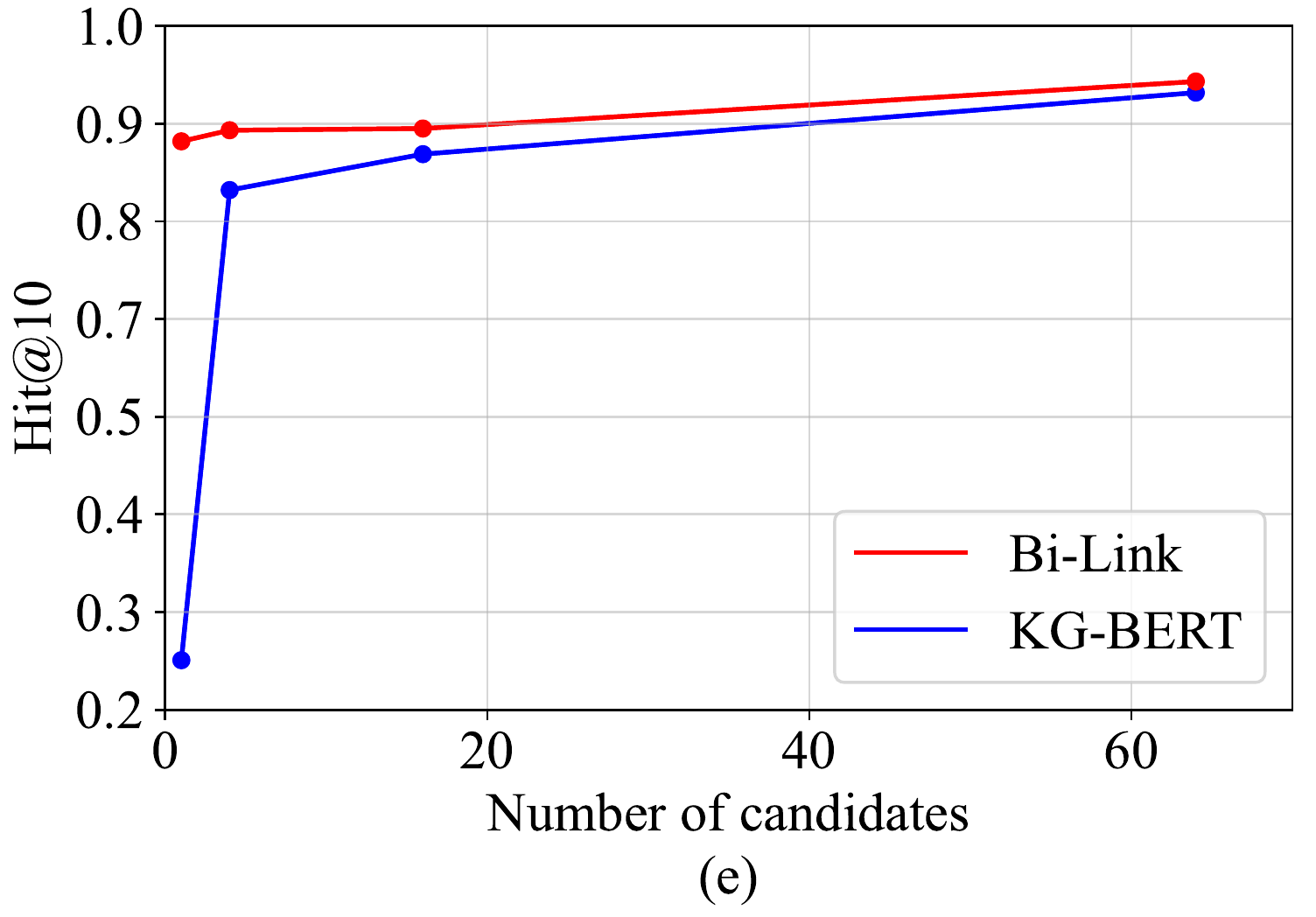}
\caption{Results of Bi-Link and KG-BERT on Zeshel and Zeshel-Ind.}
\label{fig:entity_linking}
\end{figure*}

Considering the diversity of relations, we select the size of the syntactical rule sets according to the frequency of relational grammar. In the experiments of WN18RR, we set the size of the rule set and the number of hidden dimensions to be 8, i.e., 8 pairs of reversible relational prompts. In the experiments of FB15k-237 and Wikidata5M, the size of the rule set is 32, i.e., 32 pairs of with fine-grained grammar. In particular, we use the forward relational templates in forward link prediction, and inverse relation templates in backward link prediction. During training, we use the Gaussian mixure model introduced in Section \ref{SRBP} for soft relational rule selection. At the test stage, when evaluating a single model, we use the relational template with the highest probability for the current triple $(h, r, [T])$ or $([H], r, h)$ suggesting the more frequent grammar for this relational expression. We use the average of bidirectional predictions shown in Figure \ref{fig:overview} (b). \\
We truncate entity descriptions that are longer than 64 due to computational limitation, and pad those short descriptions with neighbours’ entity name to bake in neighborhood information. 
For fair comparison with other baselines, we employ BERT~\cite{devlin2018bert} as the default pretrained language model for all link prediction methods. We initialize the bi-encoders with the weights of Hugging Face's bert-base-uncased and update them separately for fair comparison. Using better pre-trained language models as encoders is expected to improve performance further. In both training and test, embeddings extracted from the masked entity positions are zero-centered with template normalization before scoring.
Using AdamW~\cite{loshchilov2017decoupled} as the optimizer, we grid search the learning rate from \{1e-5, 2e-5, 5e-5\} with 500 warm up steps and linear decaying strategy. The weight decay parameter is 1e-4. All results are run across 3 random seeds.\\
\paragraph{Entity linking}
Entity linking ranks all candidate documents to predict a link from a named entity mention in context to its referent. Different from knowledge graph link prediction, our entity linking tasks do not provide explicit relations but candidate sets with a fixed size, as shown in Figure~\ref{fig:entity_linking}. For each mention-entity pair, the referent entity document is labelled as 1 and the negative candidates are labelled as 0. Similar with prompts for semantic textual similarity, relational prompts are not necessary to be probabilistic. For fair comparison, we fix the max sequence length to 64 in all experiments. 
% We launch intensive grid searches to investigate the effects of batch size and the number of negative candidates on supervised contrastive learning. We share the current sample's negative candidates with other samples in the same batch and call this trick negative sample sharing. Other hyperparameters are detailed in the supplementary materials.
\subsection{Metrics} Following the standard rank-based evaluation protocol of TransE ~\cite{bordes2013translating}, we evaluate the performance of our link prediction methods under the closed-world assumption (CWA). In particular, we report mean reciprocal rank (MRR) and Hits@k\space($k\in\{1,3,10\}$) of Bi-Link on four link prediction tasks mentioned above. For transductive KG link prediction, all entities within KGs are considered as candidates during evaluation. MRR is the average reciprocal rank of entities labelled in test triples. Hits@k calculates the proportion of label entities ranked among the top-k predictions. For fully inductive link prediction, we also report the ranking results among all entities. Note that our evaluation is different from that of GraIL, which randomly samples nodes as a candidate set using the probabilistic local world assumption. For entity linking, each mention-entity pair has a candidate set, which includes referent entity's document, coarse searched documents and randomly sampled in-domain documents. MRR and Hits@k are calculated among these candidate sets.
\begin{table}
\caption{Results of fully inductive link predictions.}\label{inductive}
\begin{tabular}{l cc cc}
\toprule
\multirow{2}{*}{}&\multicolumn{2}{c}{FB15k-237}&\multicolumn{2}{c}{WN18RR}\\
\cmidrule{2-3}\cmidrule{4-5}
&MRR&H@10&MRR&H@10\\
\midrule
% \multicolumn{9}{l}{\itshape embedding-based methods}\\
DKRL~\cite{xie2016representation}&3.9&7.3&6.3&11.0\\
KG-BERT~\cite{yao2019kg} & \underline{16.1}&\underline{32.1}&20.9&37.5\\
SimKGC~\cite{trouillon2016complex}&10.0&21.5&25.3&45.9\\
\midrule
Bi-Link&13.2&22.6 &\underline{27.2}&\underline{50.3}\\
Bi-Link$_{RP}$&\textbf{18.3}&\textbf{33.8}&\textbf{31.5}&\textbf{59.2}\\
\bottomrule
\end{tabular}
\end{table}

\begin{table}
\caption{Results of link predictions on Wikidata5M.}\label{wikidata}
\begin{tabular}{l cc cc}
\toprule
\multirow{2}{*}{}&\multicolumn{4}{c}{Wikidata5M-Ind}\\
\cmidrule{2-3}\cmidrule{4-5}
&MRR&H@1&H@3&H@10\\
\midrule
% \multicolumn{9}{l}{\itshape embedding-based methods}\\
DKRL~\cite{xie2016representation}&23.1&5.9&32.0&54.6\\
KEPLER~\cite{wang2021kepler}&40.2&22.2&51.4&73.0\\
BLP-ComplEx~\cite{daza2021inductive}&48.9&26.2&66.4&87.7\\
BLP-SimplEx~\cite{daza2021inductive}&49.3&28.9&63.9&86.6\\
SimKGC~\cite{wang2022simkgc}&60.1&39.4&77.7&92.4\\
\midrule
Bi-Link&\underline{60.9}&\underline{41.3} &\underline{77.9}&\underline{92.8}\\
Bi-Link$_{RP}$&\textbf{61.2}&\textbf{41.8}&\textbf{78.2}&\textbf{93.1}\\
\bottomrule
\end{tabular}
\end{table}

\subsection{Results}
\paragraph{Transductive link prediction}We report the results of transductive link prediction in Table\ref{transductive}. Our experiments show that contrastive learning can efficiently extract semantics from textual descriptions of entities and relations. With two-tailed t-tests, Bi-Link yield statistically significant improvements, demonstrating the effectiveness of bidirectional links between backward and forward textual link predictions. probabilistic rule-based prompts have shown more increases in WN18RR than in FB15k-237. We hypothesize that the parameters of the probabilistic RP model are easier to learn when relations are simple and the corresponding verbalization sounds more nature. Therefore, the model can not only learn reversible relational embeddings from reverse prompts, but also symmetric relation patterns with the proposed siamese network. \\
\paragraph{Inductive link prediction} We list the inductive link prediction results in Table \ref{inductive} and Table \ref{wikidata}. In terms of modelling entity-independent relations, we have improved the baselines by a large margin thanks to fine grained prompts. There is a more severe performance drop in SimKGC than in Bi-Link. This is probably because relations other than raw inversion are implicitly learnt through data in SimKGC, whereas Bi-Link has clear mechanism to model fine-grained inverse and symmetric relation patterns. Through our experiments, we confirm the effectiveness of template denoising designed in PromptBERT~\cite{jiang2022promptbert}, and the negative effects of self-negatives~\cite{wang2022simkgc} in the inductive link predictions.\\
\paragraph{Entity Linking}
We list the results of entity linking on Zeshel (cross-domain) and Zeshel-Ind in Table \ref{zeshel}. The first two rows show the zero-shot performance of BERT~\cite{devlin2018bert} and SpanBERT~\cite{joshi2020spanbert} followed by the performance of baseline methods and our methods. On Zeshel-Ind, Bi-Link outperforms baselines with a large margin. By contrast, Bi-Link outperforms KG-BERT by 0.3\%, indicating contrastive learning is vulnerable to domain shifts.
In Figure\ref{fig:entity_linking}, we take accuracy(Hits@1, MRR) and recall(Hits@10) as the analysis criteria. Our experiments demonstrates the importance of high-quality negative samples. With other experimental conditions being the same, using all 64 candidates gives the best results. On the in-domain entity linking task, our method outperforms KG-BERT in both linking accuracy and recall ability. \\
\paragraph{Domain Adaptation}On the cross-domain entity linking task, we are comparable to KG-BERT in accuracy, but the gap between in-domain and cross-domain zero-shot performance reveals the influence of domain shift. Our method outperforms KG-BERT in both the best results and robustness on Hits@10, meaning that the recall ability of Bi-Link is better than KG-BERT. It is worth noting that Bi-Link still perform well retain when the number of candidates is reduced, thanks to the in-batch comparison and negative candidate sharing trick. For example, in the absence of candidates, the in-domain KG-BERT has an accuracy of 50.1\% and an MRR of 0.655 on Hits@1, while the cross-domain Bi-Link has an accuracy of 50.8\% and an MRR of 0.635, demonstrating a significant superiority of our method. Our experiments show that using mention as negative samples hurts performance. \\
% with hit@10 dropping from 96.1\% to 93.6\% and hit@1 dropping from 69.7\% to 62.2\%.\\
\paragraph{Error analysis}
Running examples in Table\ref{tab:running_exp1} and \ref{tab:running_exp2} show that our method is effective . We noticed that different entities’ documents may point to the same entity, as shown in Table\ref{tab:running_exp2}. This is understandable since this dataset is community-contributed, and there may be biases during the construction of the data leading to this many-to-one situation. Under this circumstance, in the case that the documents of different entities are extremely similar, it is already difficult for people to distinguish them, which means that they can all be used as the entity’s links. To a certain extent, the discriminative ability of the model is even better than that of humans in this regard.\\
\paragraph{Limitations}
In this work, discrete prompts are manually designed by linguistic experts, which are quite expensive. When the knowledge base has a diverse collection of relations, syntax prompts may bring in too much noise to contrastive representation learning.
% \begin{figure*}
%     \centering
%     \includegraphics[width=0.98\linewidth]{results/da-hits1-mrr.pdf}
%     \caption{MRR results}
%     \label{fig:MRR_CL}
% \end{figure*}
\section{Conclusion}
In this paper, to learn a text-based model for inductive link predictions, we propose an effective contrastive learning framework, Bi-Link, with probabilistic rule-based prompts. Specifically, to create natural relational expression, the framework Bi-Link first generates rule-based prompts from syntactical patterns. We learn a Gaussian mixture model using grammatical knowledge of PLMs for syntax prediction. To better express symmetric relations, we design a symmetric link prediction model, establishing bidirectional linking between forward prediction and backward prediction. The bidirectional linking accommodates flexible self-ensemble strategies at test time. We release Zeshel-Ind dataset for in-domain zero-shot entity linking. The experiment results show that Bi-Link framework can learn robust representation via text for knowledge graph link prediction and entity linking tasks. Future work can incorporate path information into this framework to produce more explainable knowledge representations.
\bibliographystyle{ACM-Reference-Format}
\bibliography{sample-base}
\clearpage
\begin{table*}[htbp]
    \centering
        \caption{A mention its candidates from The Doctor Who}
    \label{tab:running_exp1}
    \begin{tabular}{c|l}
    \toprule
    \multicolumn{2}{c}{\textbf{The Doctor Who}}\\
    \midrule
    \multicolumn{2}{l}{\textbf{Query}:Which system has the habitable planet as \textbf{New Earth}?}\\
    \multicolumn{2}{l}{\textbf{Label}:New Earth ( Doctor Who and the Dogs of Doom ) New Earth was a planet located in the New Earth System ...}\\
    \midrule
    Method& Evidence\\
    \midrule
    % 1& &New Earth was a planet colonised by the New Earth Empire. \\
    \multirow{5}{*}{BM25} &  New Earth Republic \\
    & New Earth ( New Earth )\\
    & Tales from New Earth ( audio anthology )\\
    & Fire Island\\
    & \textbf{New Earth ( Doctor Who and the Dogs of Doom )}\\
    \midrule
    \multirow{5}{*}{Bi-Link} & New Earth ( New Earth )    \\
    & The New Earth System \\
        & \textbf{New Earth ( Doctor Who and the Dogs of Doom )} \\
            & New Earth ( Invasion of the Dinosaurs ) \\
                & New Earth ( Time of Your Life ) \\
    \bottomrule
    \end{tabular}
\end{table*}
\begin{table*}[]
    \centering
        \caption{A mention with its candidates from Sports}
    \label{tab:running_exp2}
    \begin{tabular}{c|l}
    \toprule
    \multicolumn{2}{c}{\textbf{Sports}}\\
    \midrule
    \multicolumn{2}{l}{\textbf{Query}:Which conference did \textbf{The 2009 Holiday Bowl} football game featured the 2nd pick from?}\\
    \multicolumn{2}{l}{\textbf{Label}:Pacific - 10 Conference The Pacific - 12 Conference ( Pac - 12 ) is a college athletic conference that operates in the Western United States.}\\
    \midrule
    Method& Evidence\\
    \midrule
    % 1& &New Earth was a planet colonised by the New Earth Empire. \\
    \multirow{5}{*}{BM25} & 2001 Oregon Ducks football team \\
    & Dave Hoffmann\\
    & \textbf{Pacific - 10 Conference The Pacific - 12 Conference ( Pac - 12 )}\\
    & 2003 Rose Bowl\\
    & Pacific - 12 Conference The Pacific - 12 Conference ( Pac - 12 )\\
    \midrule
    \multirow{5}{*}{Bi-Link} & \textbf{Pacific - 10 Conference The Pacific - 12 Conference ( Pac - 12 )}\\
    & Pacific - 12 Conference The Pacific - 12 Conference ( Pac - 12 )\\
        & Bowl Alliance\\
            & 2009 Holiday Bowl\\
                & Bowl Coalition\\
    \bottomrule
    \end{tabular}
\end{table*}
\end{document}